\documentclass[letterpaper, 10pt, conference]{ieeeconf}
\IEEEoverridecommandlockouts
\usepackage[pdftex]{graphicx}

\graphicspath{{../pdf/}{../jpeg/}}
\DeclareGraphicsExtensions{.pdf,.jpeg,.png}

\usepackage{amssymb} 
\usepackage{graphicx}
\usepackage{algorithm}
\usepackage{algorithmicx}
\usepackage[noend]{algpseudocode}
\usepackage{dblfloatfix}
\usepackage{siunitx}
\usepackage{arydshln}
\usepackage{graphicx}
\DeclareSIUnit\pixel{px}
\usepackage[natural]{xcolor}

\usepackage[cmex10]{amsmath}

\usepackage[pdftex]{graphicx}
\usepackage{url}
\usepackage{algorithm}
\usepackage{algorithmicx}
\usepackage[noend]{algpseudocode}
\usepackage{amssymb} 
\usepackage{graphicx}
\usepackage[natural]{xcolor}
\usepackage{changes}
\usepackage[cmex10]{amsmath}
\usepackage{mathtools}
\usepackage{url}
\usepackage{algorithm}
\usepackage{algorithmicx}
\usepackage[noend]{algpseudocode}
\usepackage{siunitx}
\usepackage{gensymb}
\usepackage{multirow}
\begin{document}
%
\title{Cyclist Trajectory Forecasts by Incorporation of Multi-View Video Information}


\author{Stefan Zernetsch, Oliver Trupp, Viktor Kress, Konrad Doll, and Bernhard Sick 
	\thanks{S. Zernetsch, O. Trupp, V. Kress, and K. Doll are with the Faculty of Engineering,
		University of Applied Sciences Aschaffenburg, Aschaffenburg, Germany
		{\tt\footnotesize stefan.zernetsch, oliver.trupp, viktor.kress, konrad.doll\{@th-ab.de\}}}
	\thanks{B. Sick is with the Intelligent Embedded Systems Lab, University of Kassel,
		Kassel, Germany
		{\tt\footnotesize bsick@uni-kassel.de}}%
}



%


\maketitle

\begin{abstract}

This article presents a novel approach to incorporate visual cues from video-data from a wide-angle stereo camera system mounted at an urban intersection into the forecast of cyclist trajectories. We extract features from image and optical flow (OF) sequences using 3D convolutional neural networks (3D-ConvNet) and combine them with features extracted from the cyclist's past trajectory to forecast future cyclist positions. By the use of additional information, we are able to improve positional accuracy by about 7.5\,\% for our test dataset and by up to 22\,\% for specific motion types compared to a method solely based on past trajectories. Furthermore, we compare the use of image sequences to the use of OF sequences as additional information, showing that OF alone leads to significant improvements in positional accuracy. By training and testing our methods using a real-world dataset recorded at a heavily frequented public intersection and evaluating the methods' runtimes, we demonstrate the applicability in real traffic scenarios. Our code and parts of our dataset are made publicly available.


\end{abstract}


%
\IEEEpeerreviewmaketitle

\section{\large Introduction}
\label{sec_introduction}
\subsection{Motivation}

In future traffic, vehicles will become more and more automated. In this scenario, automated vehicles will share the road with non-automated vehicles and vulnerable road users (VRU) such as pedestrians and cyclists. Therefore, every automated vehicle must be aware of its surroundings at all times to ensure safe interaction with other traffic participants. Automated vehicles not only have to know the current positions of other traffic participants but anticipate their movements to plan a safe trajectory. Especially the trajectories of VRU are hard to anticipate since they can change their direction quickly and are not limited to roads. Additionally, they are prone to occlusion by other road users or structures on the roadside and, therefore, hard to track by a single vehicle's sensors in complex traffic scenarios. Sensors mounted at the roadside can be used to resolve this problem. Infrastructural sensors are less susceptible to occlusions and can deliver higher positional accuracy compared to vehicle sensors. 

To anticipate the future behavior of VRU, methods that use the past VRU trajectories extracted from image data have been proposed. While these methods effectively reduce the original images' input dimensions, visual cues that can help to generate more reliable forecasts, like gaze direction, are lost. In this article, we propose a method to incorporate video information from a wide-angle stereo camera system to improve forecasts of VRU, showing that the additional information can lead to significant improvements in positional accuracy. Our method is trained and tested for cyclists, but can also be applied to pedestrians. The focus of this article is on the incorporation of multi-view video information into the forecast process. Therefore, we do not consider additional information like interaction between different road users or environmental information like maps in our proposed method or in our baseline method.

\subsection{Related Work}

Over the past years, the field of VRU trajectory forecast has become more active. Most of the proposed methods use information about the past position to forecast future positions. Goldhammer et al. extract polynomial coefficients from trajectories as input for a neural network to forecast future pedestrian trajectories~\cite{goldhammer2016phd}. The method is adapted to cyclists in~\cite{ownZernetsch_iv2016}. Compared to physical models, the neural network produces forecasts with larger positional accuracies.
To improve forecast accuracies, additional information can be added to the past trajectory. Alahi et al. propose a long short-term memory (LSTM) combined with a social pooling to add additional information about interactions with other pedestrians~\cite{Alahi_2016_CVPR}. Their method outperforms an LSTM, which uses solely past positions as input. Pool et al. use cyclists' past trajectory and add additional information about the road topology to mix specialized filters to forecast the future cyclist trajectory~\cite{Pool2017Context}. The authors are able to improve forecast accuracy by up to 20\,\% on sharp turns compared to a single model approach. Instead of a pedestrian's single past trajectory, Quintero et al. use the past trajectories of eleven body joints to provide additional information about movement and body language~\cite{quintero2018}. The joint trajectories are used in combination with balanced Gaussian process dynamical models (GPDM) to detect basic movements and forecast trajectories of pedestrians. Kress et al. use joint trajectories in combination with gated recurrent units (GRU) to forecast VRU trajectories~\cite{Kress2020}. Using the additional information from body joints, they are able to forecast more accurate positions with shorter observation periods than a solely trajectory based method. Keller and Gavrila compare the use of features derived from dense optical flow (OF) in combination with GPDM and hierarchical trajectory matching to two solely trajectory-based methods~\cite{willpedcross}. The methods based on OF improve the forecast errors by 10-50 cm at time horizons of 0-0.77 s. Sagedian et al. incorporate physical and social constraints into their forecast of pedestrian trajectories~\cite{Sadeghian2019}. They pass features extracted from a single scene image using a convolutional neural network (CNN) and hidden states of LSTM for multiple agents to an attention module. Their model outperforms baseline models like LSTM or social LSTM from~\cite{Alahi_2016_CVPR}.

Aside from intention detection in traffic scenarios, video-based human action recognition has become an active research area in recent years, with the emergence of several network architectures that can be applied to the field of VRU intention detection. In \cite{Carreira}, Carreira and Zisserman present a novel approach to video-based action recognition and compare it to the most commonly used architectures. The authors propose a two-stream approach with two 3D-ConvNets with inception architecture using an image sequence and an OF sequence as input data. The network, referred to as I3D (Inflated 3D), outperforms the other architectures on the Kinetics human action detection video dataset~\cite{Kay2017TheKH} and achieves 98.0\,\% accuracy on the UCF101 Human Action Classes dataset~\cite{Khurram}. The I3D architecture is extended to cyclist action in~\cite{ownZernetsch_icpr2020} and is used as a feature extractor in this article for cyclist trajectory forecasts. While we chose I3D due to its great performance regarding cyclist action recognition, the feature extractor used in the method is interchangeable.

\subsection{Main Contributions and Outline of this Paper}

The main contribution of this article is a method to incorporate multi-view video information from cameras mounted at a research intersection into the trajectory forecast of cyclists. By adding information from video sequences, we aim to incorporate different features, like body language or gestures into our forecast. The use of a wide-angle stereo camera system allows us to extract features of the cyclists from different view angles and resolve occlusions. We are able to improve the positional accuracy of forecasts by about 10\,\% compared to a baseline model using positional information only. For specific motion types, we are able to reduce the forecast error by up to 22\,\%. Our method is trained and tested using a dataset recorded at a public intersection showing its real-world application. Evaluation of runtimes shows that our method can be incorporated into a real-time system. The dataset and code used to train and test our final method are made publicly available \cite{zernetsch_cyclistactionrec}\cite{zernetsch_github_iv2021}.

The article's remainder is structured as follows: In Sec.~2, we describe the intersection, where the dataset is created, and the dataset itself. Sec.~3 describes the architectures of the baseline model and the video-based network, followed by the evaluation method we use. In Sec.~4, the results of both methods are visualized and compared, followed by a conclusion and a short outlook in Sec.~5.

\section{\large Test Site and Dataset}
\label{subsec:dataset}

\begin{figure}
	\begin{center}
		\vskip 0mm
		\includegraphics[width = 0.99\columnwidth, trim={0 0 0 30mm},clip]{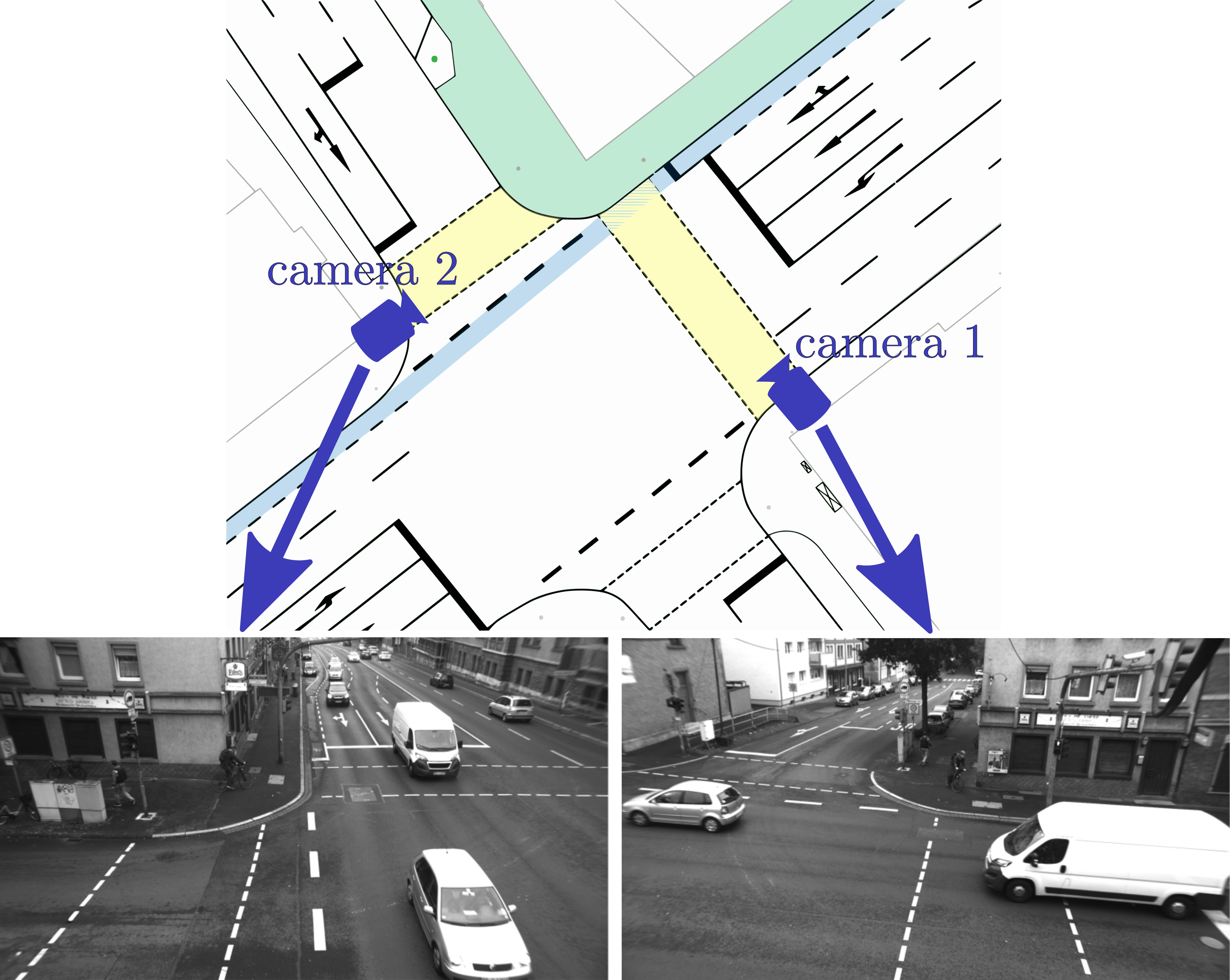}
		\vskip 0mm
		\caption{Map of intersection with sidewalk (green), bike lane (blue), and pedestrian crossings (yellow) and example images of camera 1 and 2.}
		\label{fig:intersection}
	\end{center}
	\vskip 0mm
\end{figure}

In this section, we describe the test site where the dataset
used for trajectory forecast is created, and provide a description
of the dataset itself. Except for the ground truth data, the dataset is the same as the dataset used in~\cite{ownZernetsch_icpr2020}.

The dataset was collected at an intersection close to the University of Applied Sciences Aschaffenburg using a wide-angle stereo camera system (Fig. \ref{fig:intersection})~\cite{Goldhammer}. The university's research intersection consists of four arms, two arms with five vehicle lanes each, one arm with three lanes, and one with two lanes. With up to 24,000 vehicles a day, the intersection is highly frequented. A stereo camera system, operating at 50 fps, was mounted to capture a corner of the intersection with a sidewalk (Fig. \ref{fig:intersection} green), two pedestrian crossings (Fig. \ref{fig:intersection} yellow), and a bike lane (Fig. \ref{fig:intersection} blue). The full HD grayscale cameras 1 and 2 are mounted approx. 5 m above the ground at an angle of approx. 90\degree\,relative to each other. Most occlusions by vehicles or other VRU can be avoided by using this setup. The dataset used in this work consists of samples of image sequences of cameras 1 and 2 $I_{1/2}(u, v, t)$, with $u$ and $v$ being the pixel position and $t$ being the temporal dimension and OF sequences of camera 1 and 2 $O_{1/2}(u, v, t)$. Furthermore, the set contains the past and ground truth trajectories of the cyclists $^e\mathcal{T}=\{[^ex_{t-(1s)}, ^ey_{t-(1s)}, ^ez_{t-(1s)}]...[^ex_{t}, ^ey_{t}, ^ez_{t}]\}$ and $^e\mathcal{T}_{gt}=\{[^ex_{gt;t+(0.02s)}, ^ey_{gt;t+(0.02s)}]...[^ex_{gt;t+(2.5s)}, ^ey_{gt;t+(2.5s)}]\}$ as ordered sets of positions transformed to the cyclist's ego coordinate system (indicated by $^e$) in $x$, $y$, and $z$ directions from time step $t-(1s)$ to the current time step $t$. Since we only aim to forecast the future positions in $x$ and $y$ direction, the $z$ coordinate is removed from the ground truth trajectory.

\begin{figure}
	\begin{center}
		\vskip 0mm
		\includegraphics[width = 0.99\columnwidth]{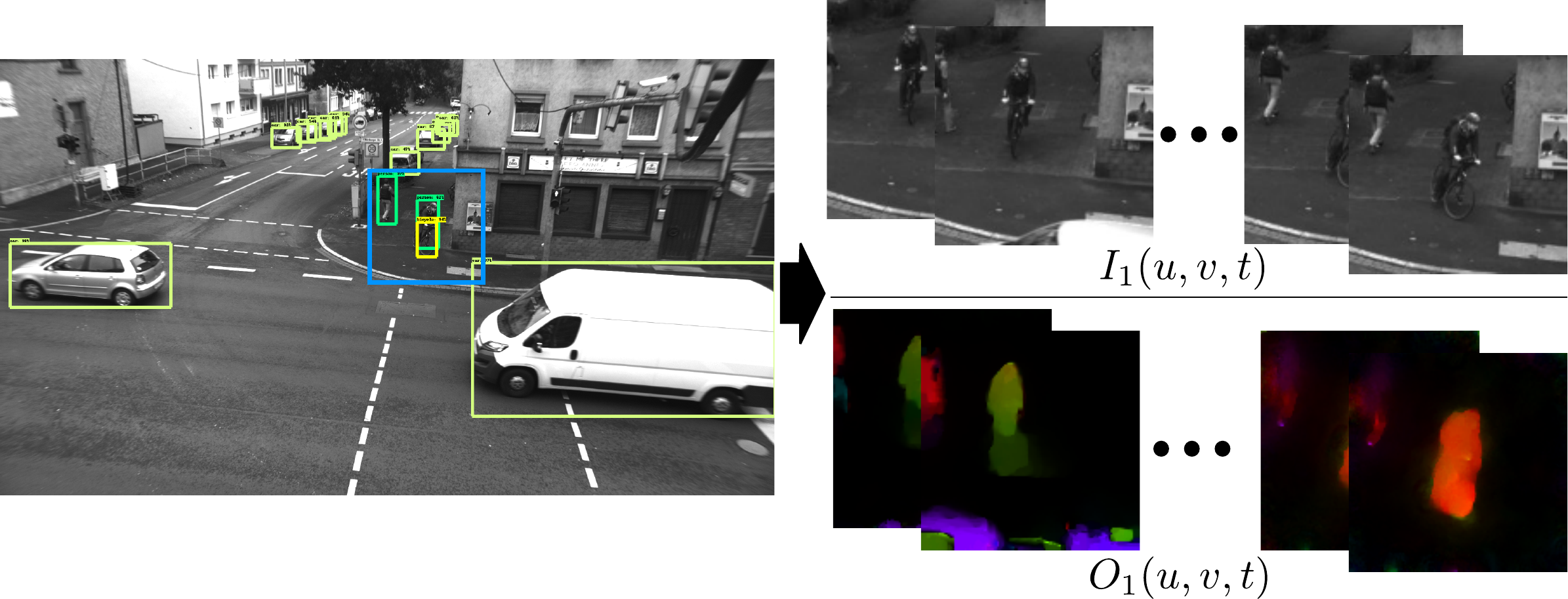}
		\vskip 0mm
		\caption{Exemplary image and OF sequence of cyclist detected on image of camera 1, with generated ROI marked in blue.}
		\label{fig:ms_creation}
	\end{center}
	\vskip 0mm
\end{figure}

Object detection is performed on each camera image using a pre-trained object detection network from \cite{Huang}, trained on the COCO dataset \cite{Lin2014} (Fig. \ref{fig:ms_creation} left), to generate image sequences. The output of the network consists of object boxes of different classes. We use detected boxes for the classes \textit{person} and \textit{bike} to create a cyclist bounding box. The detected \textit{person} boxes are associated with the \textit{bike} boxes using a simple nearest neighbor-strategy. They are combined to create a bounding box enclosing the cyclist and the bicycle. The resulting bounding box's size is enlarged by a factor $f_b$ to create a region of interest (ROI) to account for the cyclist movements from previous time steps (Fig. \ref{fig:ms_creation} left, blue box). The ROI created at time $t$ is used to extract the sequence of camera images over the last second. Instead of using all images to create the sequence, only the ones from time steps \{$t-$\SI{0.00}{\second}, $t-$\SI{0.02}{\second}, $t-$\SI{0.04}{\second}, $t-$\SI{0.06}{\second}, $t-$\SI{0.08}{\second}, $t-$\SI{0.20}{\second}, $t-$\SI{0.40}{\second}, $t-$\SI{0.60}{\second}, $t-$\SI{0.80}{\second}, $t-$\SI{1.00}{\second}\} are used. We use the five most recent images ($t-$\SI{0.00}{\second} to $t-$\SI{0.08}{\second}). Since older images become less relevant for the current movement, we chose larger gaps between the oldest five images ($t-$\SI{0.20}{\second} to $t-$\SI{1.00}{\second}) to reduce our input data. 

The length of the input horizon of \SI{1}{\second} is chosen based on previous investigations. We found that while larger input horizons lead to smaller forecast errors, the differences are not significant. An input horizon of \SI{1}{\second} has also proven to work well for pedestrian action recognition \cite{goldhammer2016phd}. The extracted images are resized to $192\times192$ px and stacked to a three-dimensional array resulting in an image sequence $I_{1/2}(u, v, t)$ with dimensions $192\times192\times10$ px, the third dimension being the temporal dimension. This size is chosen because it corresponds to approximately the size of an ROI in the region where most cyclists occur.

The OF sequence is created from the image sequence using the pre-trained OF network from \cite{Sun}. For each of the nine consecutive image pairs in $I_{1/2}(u, v, t)$, an OF image with two channels (OF in $u$ and $v$ direction) is created and stacked along the channel axis resulting in an optical flow sequence $O_{1/2}(u, v, t)$ with dimension $192\times192\times18$ px.

The trajectories are created by tracking the head position of the cyclist in each camera image. To ensure that the correct cyclist in every scene is tracked, this process is performed semi-automatically and corrected manually if needed. Occlusions of a cyclist in camera images are handled by tracking the position using a Kalman Filter with a constant velocity model. The 3D head position in world coordinates $^w\vec{p}_t=[^wx_{t}, ^wy_{t}, ^wz_{t}]$ for every time step $t$ is calculated using triangulation. As network input, we use the trajectory of the past second in ego coordinates, which we achieve by transformation in a way that the origin of the coordinate system is the head position of the cyclist at the current time and the $y$ axis parallel to the cyclist's longitudinal movement direction. The OF sequences and trajectories are made publicly available \cite{zernetsch_cyclistactionrec}. Due to German privacy laws, we are not permitted to publish image sequences.


\section{\large Method}
\label{subsec:basic_movement_detection}
This section describes our overall method. Sec. \ref{subsec:baseline} shows our baseline method, which uses GRU to forecast trajectories based on solely past positions. In Sec. \ref{subsec:trajectory_forecast}, we describe how we incorporate video information into the forecast. The evaluation method is described in Sec. \ref{subsec:evaluation_methods}.

\subsection{Baseline Method}
\label{subsec:baseline}

\begin{figure}
	\begin{center}
		\includegraphics[width = 0.8\columnwidth]{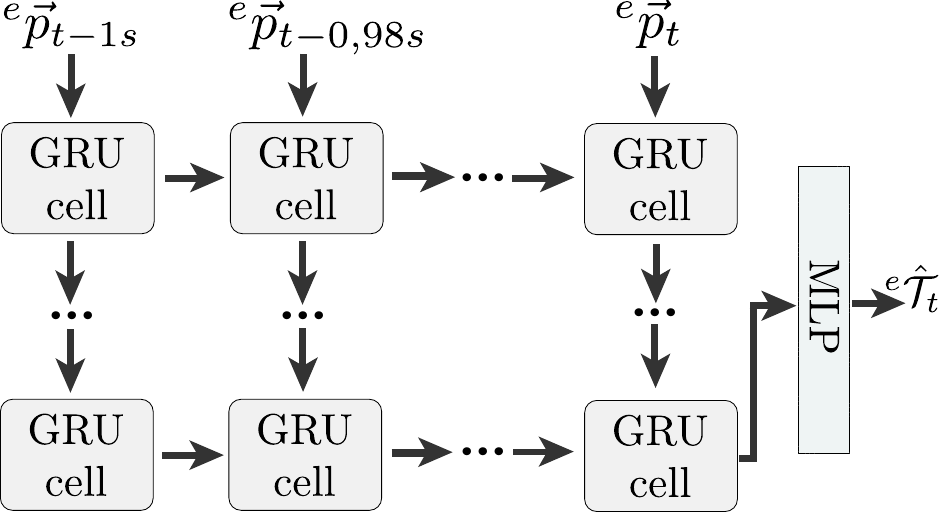}
		\caption{Architecture of the GRU baseline network.}
		\label{fig:gru_arch}
	\end{center}
	\vskip 0mm
\end{figure}

We use a GRU, which uses solely positional information as input data, as a baseline method to compare against our proposed algorithm. In preliminary tests, we compared the use of MLP, LSTM, and GRU against each other, where the smallest positional errors were achieved by GRU. Fig. \ref{fig:gru_arch} depicts the architecture of our baseline model. The positions of the last second are fed to GRU cells. The number of hidden GRU cell layers and hidden output states are part of a hyperparameter optimization. The GRU cell's output in the final layer is passed to an MLP architecture with linear output. The number of hidden layers and neurons of the MLP are also part of the hyperparameter optimization. As loss function, we use

\begin{equation}
\begin{split}
L_{t}(^e\mathcal{T}_{gt;t}, &^e\hat{\mathcal{T}}_{t})=\\
\frac{1}{|\mathcal{H}|}\sum_{h\in\mathcal{H}}&\frac{(^ex_{gt;t+h}-^e\hat{x}_{t+h})^2 + (^ey_{gt;t+h}-^e\hat{y}_{t+h})^2}{h},
\end{split}
\label{eq:loss_traj}
\end{equation}
with the ground truth and forecasted trajectory $^e\mathcal{T}_{gt;t}$ and $^e\hat{\mathcal{T}}_{t}$ in ego coordinates and the set of all forecasted time horizons $\mathcal{H}=\{\SI{0.02}{\second}, \SI{0.04}{\second}, ..., \SI{2.5}{\second}\}$. Through division by $h$, we achieve a normalization of the quadratic forecast error to its individual forecast horizon. The model is implemented in TensorFlow \cite{tensorflow2015} and trained using the adaptive moment estimation (Adam) optimizer.

\subsection{Video-Based Trajectory Forecast}
\label{subsec:trajectory_forecast}

\begin{figure}
	\begin{center}
		\includegraphics[width = 0.9\columnwidth]{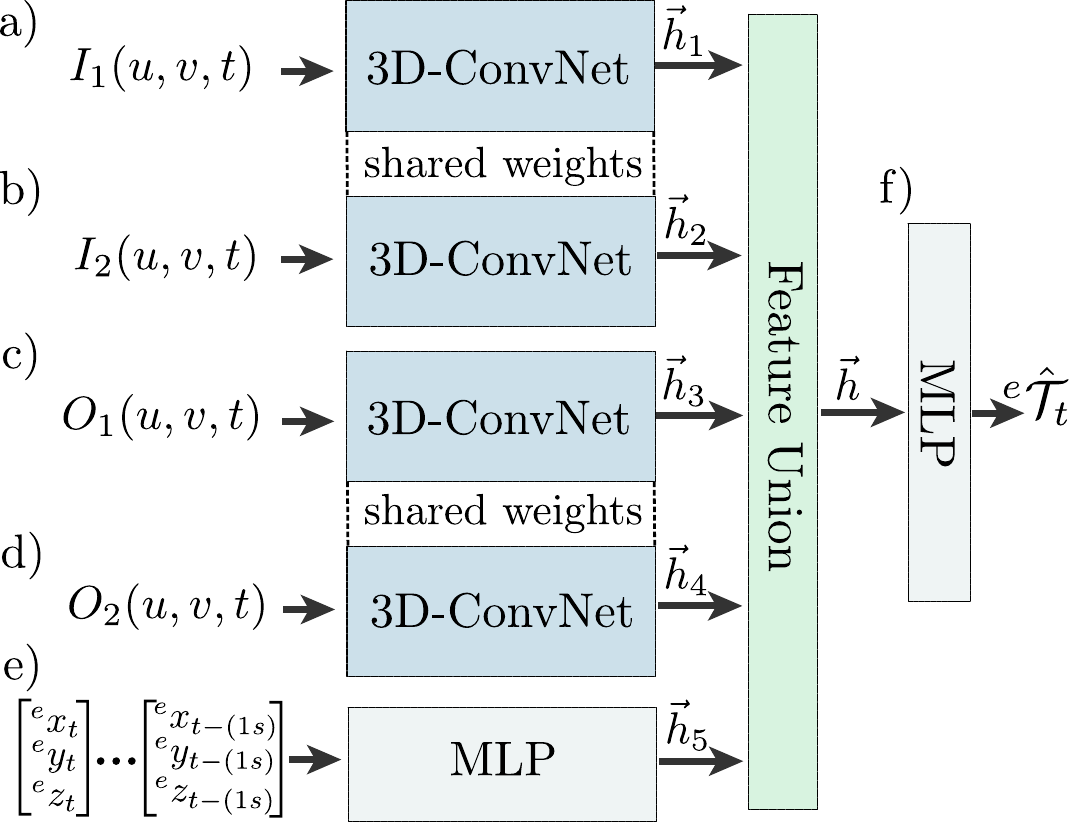}
		\vskip 0mm
		\caption{Motion sequence network architecture.}
		\label{fig:bmd_arch}
	\end{center}
	\vskip 0mm
\end{figure}

To incorporate video information into our trajectory forecast process, we implement the I3D architecture from \cite{Carreira} to extract features from our image and OF sequences. We extended the architecture to the wide-angle stereo camera system described in Sec.~\ref{subsec:dataset}. The extracted video features are fused with features extracted from the input trajectory using a feature union. Our architecture, which we call MS-Net (motion sequence), is depicted in Fig.~\ref{fig:bmd_arch}, with one Two-Stream 3D-ConvNet per camera (Fig.~\ref{fig:bmd_arch}a-d) using the respective image sequence $I_{1/2}(u, v, t)$ and OF sequence $O_{1/2}(u, v, t)$ as input. We pass the trajectory of the past second $^e\mathcal{T}_{t}$ to an MLP architecture (Fig.~\ref{fig:bmd_arch}e). The outputs of the Two-Stream 3D-ConvNets and the MLP are concatenated and passed to an MLP architecture with linear output, which outputs the trajectory forecast (Fig.~\ref{fig:bmd_arch}f).

\begin{figure}
	\begin{center}
		\vskip 0mm
		\includegraphics[width = 0.6\columnwidth]{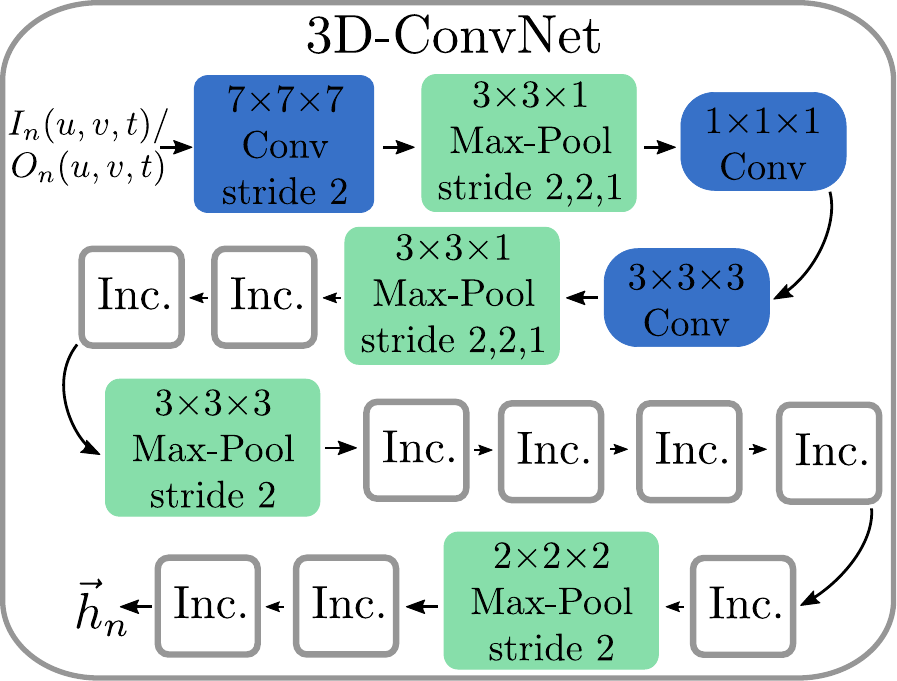}
		\vskip 0mm
		\rule{0.51\columnwidth}{1pt}
		\vskip 1mm
		\includegraphics[width = 0.6\columnwidth]{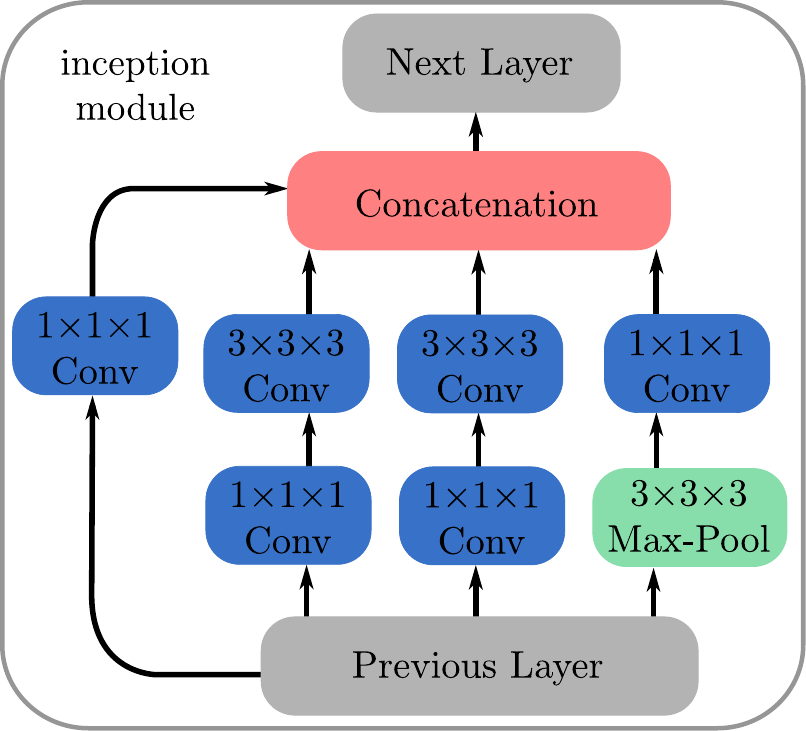}
		\vskip -1mm
		\caption{3D-ConvNet architecture (top) and detailed inception module (Inc., bottom) with 3D convolution and max-pool layers with dimension width$\times$height$\times$temporal length.}
		\label{fig:i3d}
	\end{center}
	\vskip 0mm
\end{figure}

The architecture of the 3D-ConvNets is depicted in Fig.~\ref{fig:i3d}. The top figure shows the overall architecture, and the bottom figure shows a detailed description of the I3D inception modules. Compared to the original implementation \cite{Carreira}, we used similar strides in pooling and convolution layers. However, the authors used 64 RGB and 64 OF images with a resolution of 224$\times$224 px in an input sequence, covering a temporal window of \SI{2.56}{\second}. In contrast, we only use ten grayscale frames in the input image sequence and nine two-channel OF frames with a resolution of 192$\times$192 px, covering a temporal window of \SI{1}{\second}.

To train the network, we use the same loss as for the baseline network (Eq.~\ref{eq:loss_traj}) in combination with the Adam optimizer. The weights of the 3D-ConvNets are shared between the networks of the image sequences (Fig.~\ref{fig:bmd_arch}a/b) and OF sequences (Fig.~\ref{fig:bmd_arch}c/d) of cameras 1 and 2, since these networks have similar inputs and need to extract similar features. The network was implemented using TensorFlow~\cite{tensorflow2015}.

\subsection{Evaluation Methods}
\label{subsec:evaluation_methods}

We evaluate the benefit of incorporating video information into the forecast by comparing the baseline method's positional error with MS-Net's positional error. We first calculate the average Euclidean errors (AEE) for every forecast horizon $h\in\mathcal{H}$ by
\begin{equation}
\begin{split}
AEE_{h}=\frac{1}{N_s}\sum_{i=1}^{N_s}||^e\vec{p}_{gt;h;i}-^e\hat{p}_{h;i}||_2
\end{split}
\label{eq:aee}
\end{equation} with the number of samples $N_s$ and $^e\vec{p}_{gt;h;i}$ and $^e\hat{p}_{h;i}$ as ground truth and forecasted positions of the $i$-th sample at forecast horizon $h$. Using Eq. \ref{eq:aee}, we get the average distance between forecasted and ground truth position for every forecast horizon, which we combine into the average specific AEE (ASAEE) from \cite{goldhammer2016phd}. Eq. \ref{eq:asaee} shows how the ASAEE is calculated. 

\begin{equation}
ASAEE=\frac{1}{|\mathcal{H}|}\sum_{h\in\mathcal{H}}\frac{AEE_h}{h},
\label{eq:asaee}
\end{equation}
We receive a single score consisting of the average AEE normalized according to their respective forecast horizon, which we use to compare our methods' overall performances.

In addition to using all video input sources combined, we investigate the use of image and optical flow sequences separately. We call the models MS$_{I}$, MS$_{OF}$, and MS$_{I;OF}$, where the index describes which inputs ($I$ image sequences, $OF$ optical flow sequences) are used in addition to the trajectory. Models where only image or OF sequences from camera 1 or 2 are used are referred to as MS$_{I;1}$, MS$_{I;2}$, MS$_{OF;1}$, and MS$_{OF;2}$. For the best models, we also create the results for different movement types, to evaluate if there are certain scenarios where incorporation of video information is especially useful. We identify the motion states \textit{wait}, \textit{start}, \textit{stop}, \textit{move straight}, \textit{turn left}, and \textit{turn right}. Due to traffic lights, about half of our samples are \textit{wait} samples. Since \textit{wait} forecasts show much lower forecast errors than the other motion states, we also create the results for all samples except \textit{wait} (denoted by $\overline{\text{wait}}$).

To evaluate whether any differences between the methods' results are significant or merely exist due to chance, we perform a statistical test. Therefore, we divide our dataset into sub-datasets, which we achieve by grouping all samples of one VRU to a subset. We create the ASAEE for every subset and every model and determine ranks for every subset. The ranks are used to perform the Friedman-test to determine whether significant differences in the ranks exist~\cite{friedman}. If the differences prove significant, we perform the Nemenyi-test to determine which of the models show significant differences~\cite{nemenyi}.
\section{Experimental results}
\label{sec_ResultsOutline}

\subsection{Training the networks}

\begin{figure*}
	\begin{center}
		\includegraphics[width = 1.85\columnwidth]{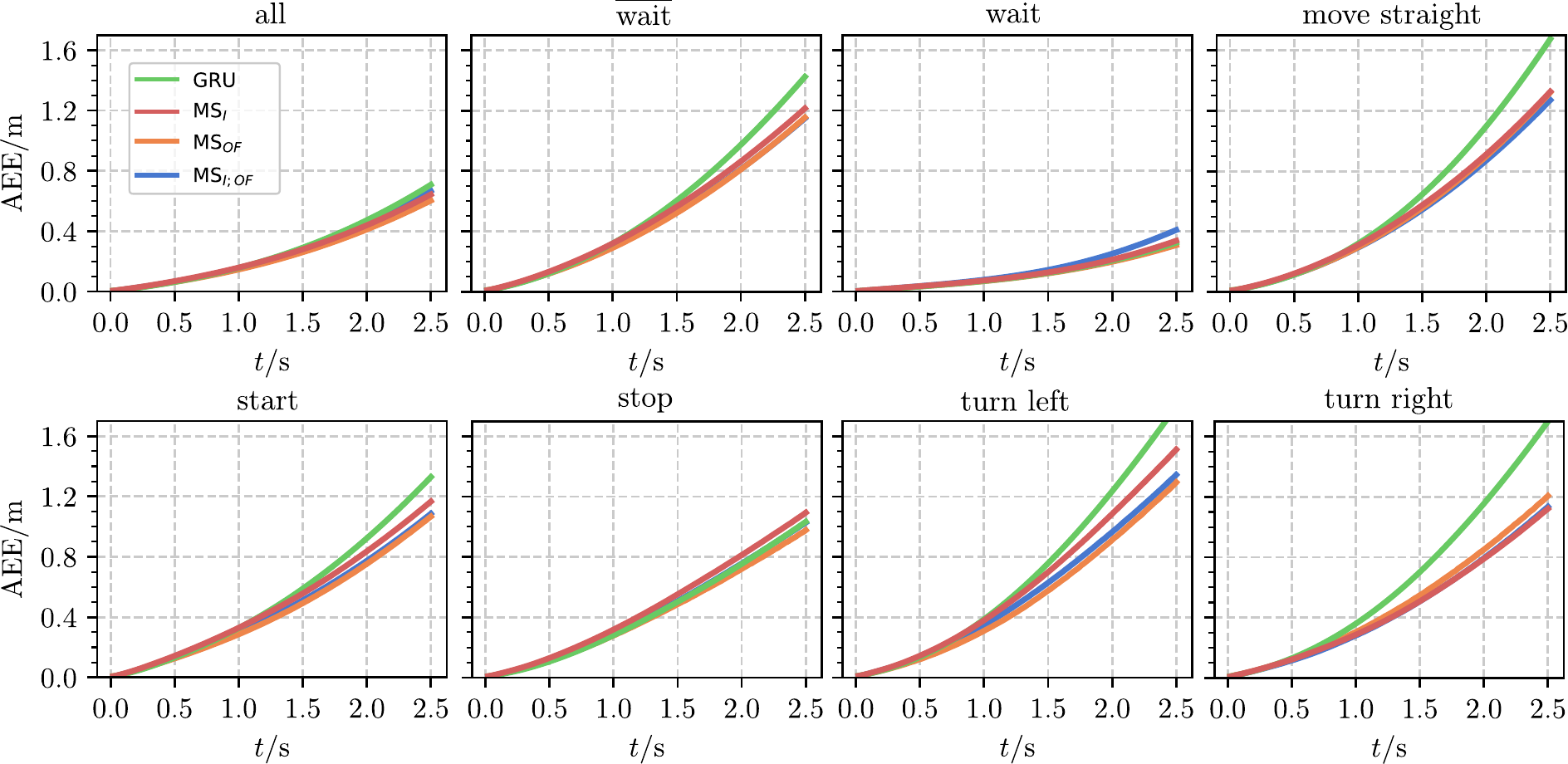}
		\caption{Comparison of AEE over forecast horizon of GRU (green), MS$_{I}$ (red), MS$_{OF}$ (orange), and MS$_{I;OF}$ (blue) for different motion states.}
		\label{fig:aee}
	\end{center}
	\vskip 0mm
\end{figure*}

\begin{figure}
	\begin{center}
		\includegraphics[width = 0.99\columnwidth]{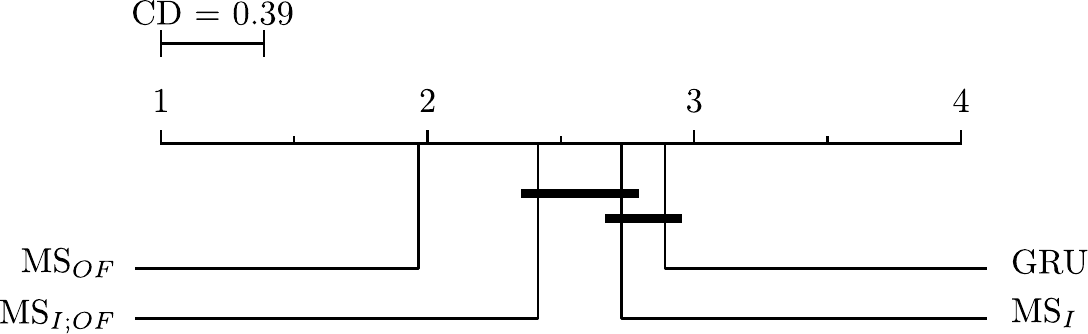}
		\vskip 0mm
		\caption{Performance comparison of different models by mean ranks. Groups of models that show no significant differences are connected ($p=0.05$).}
		\label{fig:nem}
	\end{center}
\end{figure}

\subsubsection*{Baseline}
In the first step, we perform a hyperparameter search for the GRU model. The network parameters we optimize are the number of hidden cells, the size of the hidden state of the GRU cells, and the number of hidden layers and neurons of the MLP. Additionally, we train the network using different batch sizes and learning rates. All possible combinations are trained and evaluated using the training and validation set. The model with the smallest validation error is used to create the test results. Table~\ref{tab:param_sweep_rnn} shows the parameters we investigated and the best configuration. The number of hidden states per layer are represented as vectors, where the elements represent the number of hidden states in consecutive layers.

\subsubsection*{MS-Net}

Since the MS-Net training takes considerably longer (about two weeks vs. about 2 hours for the GRU), we did not perform an extensive parameter sweep. By variation of the number of CNN filters in the I3D ConvNets, we found that we can achieve similar results with only 20\,\% of feature maps produced by the conv layers compared to the original implementation, which leads to shorter training and inference times. Aside from the number of filters, we use the network configuration from the original implementation. The MLP for feature extraction from the trajectory and the output MLP both consist of three hidden layers, with 100 neurons each. The networks are trained until the best validation score is reached and used to create the individual test results.

\subsection{Test Results}

\begin{table} [b]
	\begin{center}
		\caption{Evaluated combinations of different inputs, with image ($I$) and optical flow ($OF$) sequences of cameras $1$, $2$, or both, and the baseline model GRU with only trajectory ($T$) input.}
		\label{tab:overall_asaee_results}
		\begin{tabular}{|c | c | c | c | c | c | c |}
			\hline
			Modellname & $I1$ & $I2$ & $OF1$ & $OF2$ & $T$ & ASAEE/(m/s)\\ \hline
			GRU &  &  &  &  & x & 0,211 \\ \hline
			MS$_{I;1}$ & x &  &  &  & x & 0,209 \\ \hline
			MS$_{I;2}$ &  & x &  &  & x & 0,212 \\ \hline
			MS$_{OF;1}$ &  &  & x &  & x & 0,198 \\ \hline
			MS$_{OF;2}$ &  &  &  & x & x & 0,197 \\ \hline
			MS$_{I}$ & x & x &  &  & x & 0,196 \\ \hline
			MS$_{OF}$ &  &  & x & x & x & 0,189 \\ \hline
			MS$_{I;OF}$ & x & x & x & x & x & 0,194 \\ \hline
		\end{tabular}
	\end{center}
	
\end{table}

\begin{table}
	\caption{GRU hyperparameter sweep. The combination with the best score is marked in blue.}
	\vskip -8mm
	\begin{center}
		\begin{tabular}{|c | l|}
			\hline
			GRU layers& [50], {\color{blue}[50 50]}, [100], [100 100], \\ \hline
			MLP layers& [10 10], {\color{blue}[100 100]}, [100 100 100], [1000 1000], \\ \hline
			Batch size&1000, {\color{blue}5000}, 10000 \\ \hline
			learning rate &$5\cdot10^{-4}$, {\color{blue}$10^{-4}$}, $10^{-5}$ \\ \hline
		\end{tabular}
	\end{center}	
	\label{tab:param_sweep_rnn}
	\vskip 0mm
\end{table}

\begin{table} [b]
\begin{center}
	\caption{Mean ASAEE values of VRU sets of all models for different motion types.}
	\vskip 0mm
	\scalebox{0.8}{
		\begin{tabular}{| l | c | c | c | c | c | c | c | c |}
			\hline
			&All& $\overline{\text{wait}}$ & wait & straight & start & stop & left & right  \\ \hline	
			GRU& 0.290 & 0.395 & 0.110 & 0.414 & 0.383 & 0.292 & 0.456 & 0.478 \\ \hline
			MS$_{I}$& 0.277 & 0.377 & 0.117 & 0.378 & 0.372 & 0.344 & 0.423 & 0.379 \\ \hline
			MS$_{OF}$& 0.252 & 0.342 & 0.109 & 0.359 & 0.333 & 0.277 & 0.357 & 0.400 \\ \hline
			MS$_{I;OF}$& 0.268 & 0.357 & 0.121 & 0.360 & 0.347 & 0.310 & 0.378 & 0.408 \\ \hline
		\end{tabular}}
		\label{tab:asaee_results}
	\end{center}
	\vskip 0mm
\end{table}

We first compare the overall ASAEE of every model in Table~\ref{tab:overall_asaee_results}. We can see, that the addition of image sequences of only one camera (MS$_{I;1}$ and MS$_{I;2}$) leads to similar errors compared to the baseline model. By combining the two camera inputs (MS$_{I}$), we are able to improve the forecast error by 7\,\%. By using a single optical flow sequences as input (MS$_{OF;1}$ and MS$_{OF;2}$), we already achieve overall ASAEE which are 6\,\% lower than the baseline model's. By combining the two inputs (MS$_{OF}$), we achieve a 10\,\% smaller ASAEE than the baseline model. In both cases, the combination of both camera sources lead to an improved result. We see two main reasons for this. First, the use of both cameras helps to resolve occlusions that can occur within one camera. Second, depending on the current motion state, a certain view of the cyclist can help to gain better understanding of the cyclist's intention. E.\,g., the starting intention of a waiting cyclist can be forecasted more reliable using an image sequence that show the cyclist from the side, while the view from the front or the back is better used for the forecast of turning motions. The combination of image and OF sequences produces only slightly better results compared to image sequences only and worse results compared to OF sequences only.

\begin{figure}
	\begin{center}
		\includegraphics[width = 0.99\columnwidth]{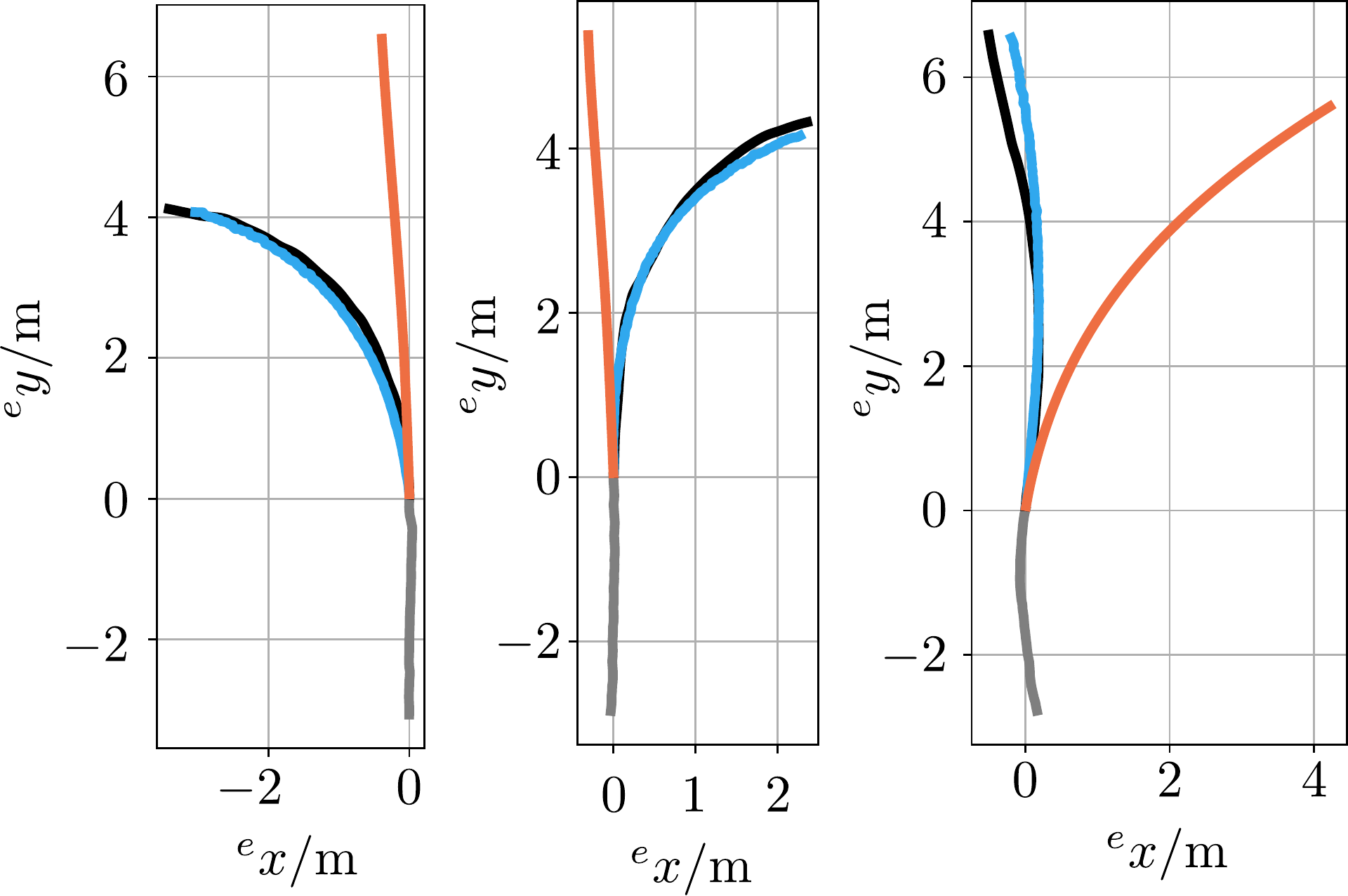}
		\caption{Examples of \textit{turn left}, \textit{move straight}, and \textit{turn right} forecasts, with input trajectories (gray), ground truth trajectories (black), and trajectory forecasts by MS$_{OF}$ (blue) and GRU (orange).}
		\label{fig:example_forecast}
	\end{center}
	\vskip 0mm
\end{figure}

For each best model regarding their input source (MS$_I$, MS$_{OF}$, MS$_{I;OF}$, and GRU), we create the results for all VRU subsets to perform statistical tests. We also evaluate the models regarding specific motion types as described in Sec.~\ref{subsec:evaluation_methods}. Fig~\ref{fig:aee} and Table~\ref{tab:asaee_results} show the overal AEE over the forecasted horizons and the mean ASAEE values of the subset split by VRU. Fig.~\ref{fig:nem} shows the results of the Nemenyi-Test for all motion types based on the VRU specific ASAEE, where models connected by a black line do not show significant differences. The model with OF sequences as additional information generates the lowest mean ASAEE, followed by the model with the combination of image and OF sequences. The model with image sequences produces lower ASAEE compared to the baseline model. However, the test shows no significant difference between MS$_{I}$ and GRU. MS$_{OF}$ shows significantly lower ASAEE compared to MS$_{I;OF}$. As for the results of different movements, MS$_{I;OF}$ only performs significantly worse than MS$_{OF}$ for waiting samples. For the rest of the movement types, there is no significant difference. Aside from \textit{turn right} and \textit{move straight}, there is no significant difference between the performance of MS$_{I}$ and GRU. One reason for the OF sequences to lead to better performance than image sequences might be because they already encode the magnitude and direction of the cyclists' motions, which is arguably an important feature to forecast movements. If we look at the motion state \textit{wait}, where almost no cyclist movements occur, we see no significant improvement by MS$_{OF}$ compared to GRU. Another reason might be that the image sequences contain a lot of information that is not considered useful for the task, like background objects.

When we look at the different motion types, we see that the most considerable improvements are achieved for the motion states \textit{turn left}, \textit{turn right}, and \textit{move straight}. By looking at the samples with the most considerable improvements compared to the baseline models, we find that, in the case of turns, the baseline model often forecasts a straight movement at the beginning of a turn, while the MS-Net forecasts a turn much closer to the real trajectory. In the case of \textit{move straight}, the baseline model often forecasts a turn due to lateral oscillation in the past cyclist trajectory. If we look at the video sequences of these cases, we can see that the cyclists start turning their head early into the direction they will turn to before noticeable bends in the cyclists' past trajectories occur. In some cases, the cyclists stop pedaling before turning. Fig.~\ref{fig:example_forecast} shows three examples of forecasts by MS$_{OF}$ and GRU compared to the ground truth. Another motion type where we see a large improvement by adding video information is \textit{start}. While the baseline model often over or underestimates the distance the cyclist will travel, MS-Net achieve much more accurate starting forecasts.

To assess whether we can use our developed methods in a real-time system, we measure the different methods' runtimes.  The runtime of the baseline method is below 1 ms. The runtimes of MS$_{I}$, MS$_{OF}$, and MS$_{I;OF}$ are measured at 10~ms, 13~ms, and 20~ms, using an NVIDIA Titan V GPU. With further optimization and current hardware developments, we are confident that our method can be implemented into a real-time system.

\section{\large Conclusions and Future Work}
\label{sec_conclusion}

In this article, we presented an approach to incorporate video information of a wide-angle stereo camera system at an urban intersection into the forecast of cyclist trajectories. By combining the sequences from the two cameras, we were able to resolve occlusions and extract more appropriate features for the forecast of certain movement types by utilizing different view angles. Our evaluations showed that especially the use of OF sequences leads to a significant improvement regarding forecast errors. The overall forecast error was reduced by 7.5\,\%. The most considerable improvements were achieved for the motion states \textit{turn left}, \textit{turn right}, and \textit{move straight}, with the largest gain of 22\,\% for \textit{turn left}. We attribute the improvements to the MS-Net being able to identify early movement indicators like turning of the head towards the movement direction. Our runtime evaluations show that our method can be implemented into a real-time system. Compared to the OF based approach, the model using image sequences generated no significant improvements in most cases. We attribute this to the large amount of information within the image sequences. Therefore, in our future work, we plan to incorporate attention mechanisms into our approach to reduce the information to relevant parts of the sequence. Since the goal of this article was to demonstrate the gain of incorporating multi-view image information into the forecast process, we chose rather simplistic forecast methods. In our future work, we plan to combine our approach with more sophisticated methods that incorporate social and physical constraints~\cite{Sadeghian2019}. Furthermore, we will compare our method to models based on body joints, especially whether our approach can reduce the needed input length~\cite{Kress2020}. We also plan to extend our approach from a deterministic forecast to a probabilistic forecast, to evaluate whether our approach can create more reliable forecasts than an approach solely based on trajectories~\cite{ownZernetsch_iv2019}.

\section{\large Acknowledgment}

This work results from the project DeCoInt$^2$, supported by the German Research Foundation (DFG) within the priority program SPP 1835: ``Kooperativ interagierende Automobile'', grant numbers DO 1186/1-2 and SI 674/11-2. Additionally, the work is supported by ``Zentrum Digitalisierung Bayern''.






\bibliographystyle{IEEEtran}
%
{\small
	\bibliography{IEEEabrv,sz}
}

\end{document}